\documentclass[10pt,twocolumn]{article}

\usepackage{cvpr}
\usepackage{times}
\usepackage{epsfig}
\usepackage{graphicx}
\usepackage{amsmath}
\usepackage{amssymb}
\usepackage{siunitx}
\usepackage{comment}
\usepackage{tabularx}
\usepackage{tabu}
\usepackage{subfig}
\usepackage{MnSymbol}

\usepackage{tikz}
\usepackage{pgf-pie}

\usepackage{pgf,pgfplots}
\pgfplotsset{compat=1.14}
\usepackage{mathrsfs}
\usetikzlibrary{arrows}
\usepackage{tkz-graph}
\usetikzlibrary{arrows.meta}

\usepackage{pst-all}
\usepackage{pst-solides3d}

\newcommand{\db}[1]{{\color[rgb]{1,0,0} {\Large #1}}}

\usepackage[pagebackref=true,breaklinks=true,letterpaper=true,colorlinks,bookmarks=false]{hyperref}

\cvprfinalcopy 

\def\cvprPaperID{7833} 

\ifcvprfinal\fi
\begin{document}

\title{Cameras Viewing Cameras Geometry}

\author{Danail Brezov\\ University of Architecture, Civil Engineering and Geodesy (UACEG)\\Sofia, Bulgaria\\
{\tt\small danail.brezov@gmail.com}
\and
Michael Werman\\
The Hebrew University\\
Jerusalem, Israel\\
{\tt\small michael.werman@mail.huji.ac.il}
}

\maketitle

\begin{abstract}
A basic problem in computer vision is to understand the structure of a real-world scene given several images of it.
Here we study several theoretical aspects of the intra multi-view geometry of calibrated cameras  when all that they can reliably recognize is each other.   With the proliferation of wearable cameras, autonomous vehicles and drones, the geometry of these multiple cameras is  a timely and relevant problem to study. 
\end{abstract}

\section{Introduction}

A basic problem in computer vision is to understand the structure of a real-world scene given several images of it.
This goes under the rubric of multi-view geometry or SLAM, simultaneous localization and mapping.
With the proliferation of wearable cameras, autonomous vehicles and drones, the geometry of these multiple cameras is  a timely and relevant problem to study. 
Here we study several theoretical aspects of the intra multi-view geometry of calibrated cameras  when all that they can reliably recognize is each other.

We treat both the  general 3D case as well as the restricted 2D setup as it is often an adequate,  simpler and more robust model for people and vehicles restricted to a planar surface.

Previous work includes using the images of other cameras to help
 reduce the number of required corresponding points to compute  epipolar geometry, \cite{8451727,satorecovering} and the vast literature on multiview geometry and pose estimation.


\begin{figure}
    \centering
    \subfloat[]{%
  \begin{minipage}{0.62\columnwidth}
  \includegraphics[width=\columnwidth]{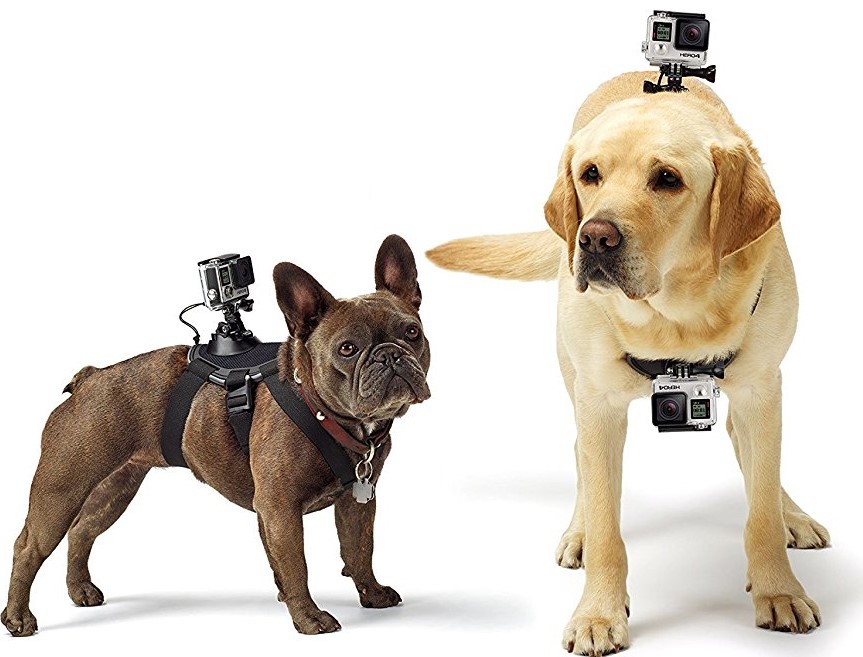}\\
   \includegraphics[width=\columnwidth]{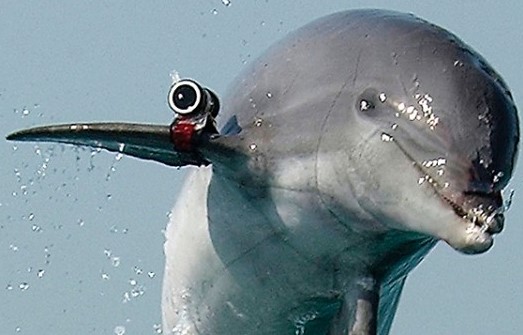}\\
  \end{minipage}%
  }
  \begin{minipage}{0.38\columnwidth}
  \includegraphics[width=0.48\columnwidth]{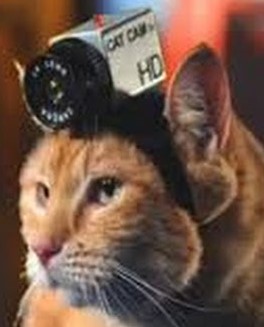}
        \includegraphics[width=.48\columnwidth]{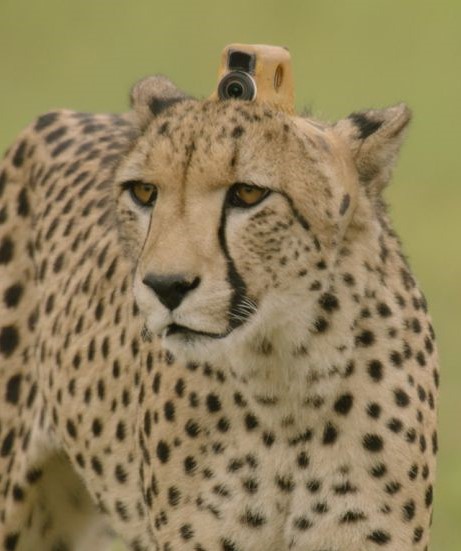}
  \includegraphics[width=\columnwidth]{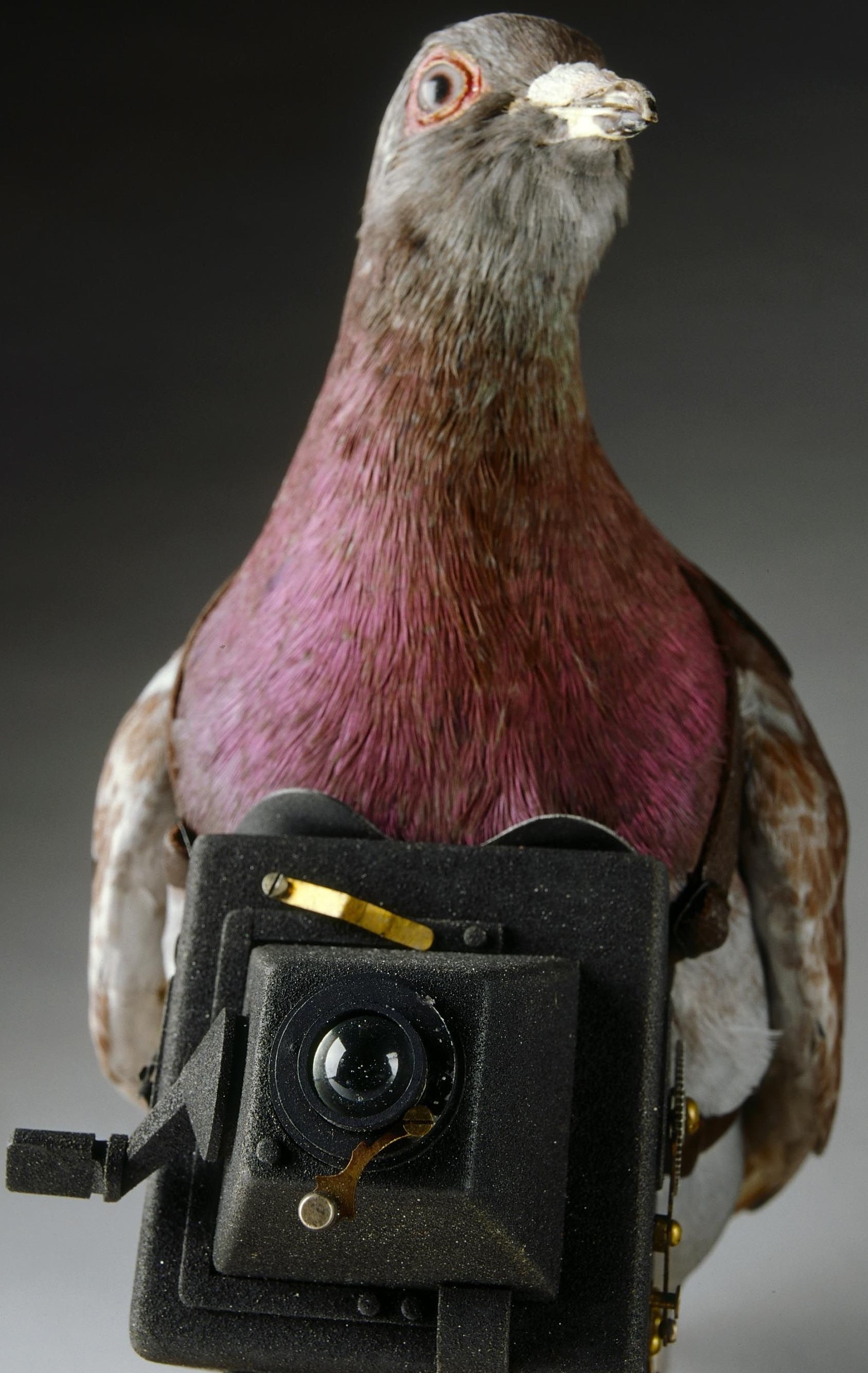}
\end{minipage}%
     \caption{Wearable cameras are ubiquitous}
    \label{fig:my_label}
\end{figure}

\section{Setup}
Let there be $n$ calibrated
cameras parametrized by their external parameters; $R_i,T_i$.

Camera $i$ sees some subset of the others as pixels, camera $i$ sees camera $j$,
$i \rightarrow j$, in homogeneous coordinates as:  
$$
[R_i |-R_i T_i] \begin{bmatrix}T_j\\1\end{bmatrix}=R_i(T_j-T_i)
$$ 

\begin{figure}
   \centering
   \includegraphics[width=\columnwidth]{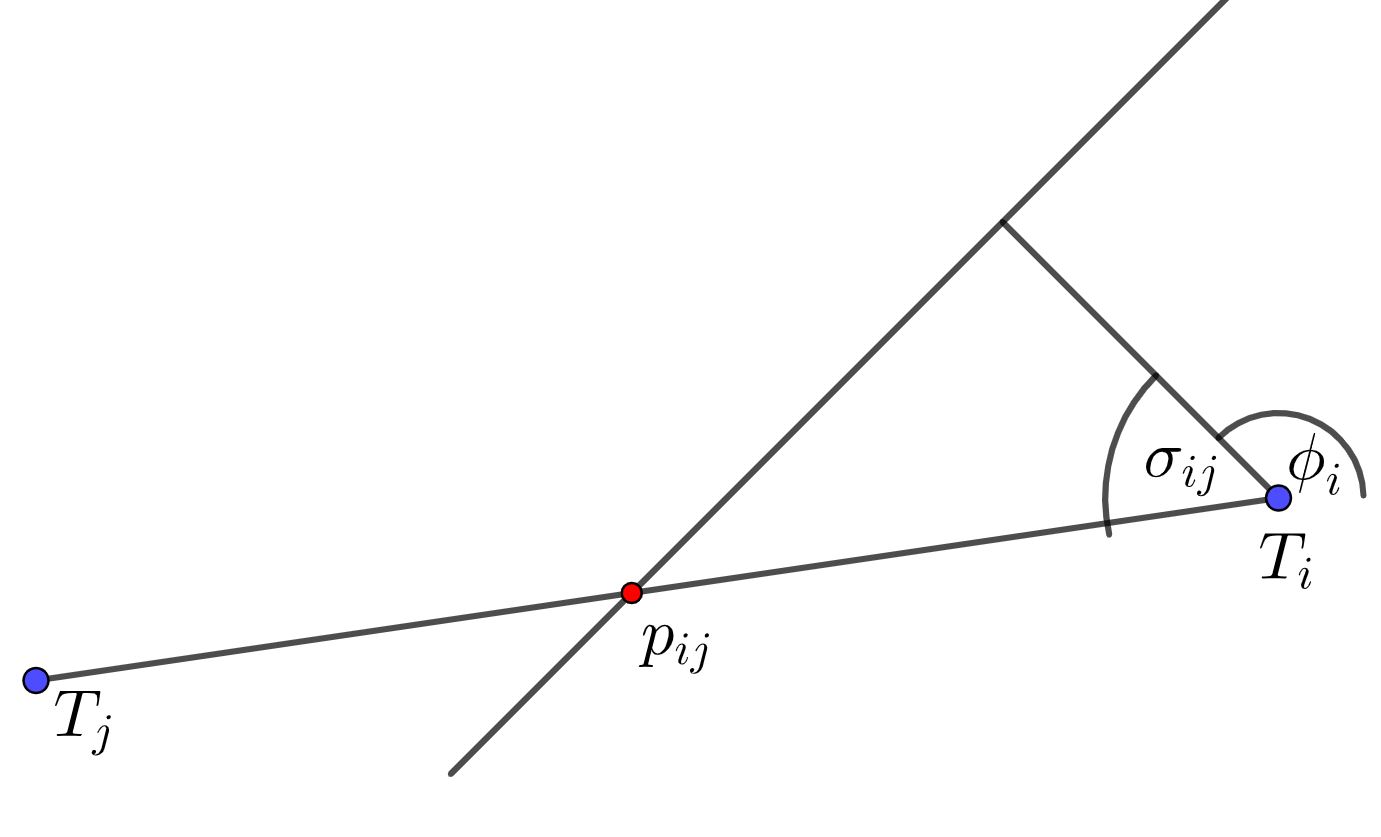}
\caption{The camera at $T_i$ with orientation $\phi_i$ sees the camera at $T_j$ at pixel $p_{ij}$, which is $\sigma_{ij}$ radians off $\phi_i$}
\label{2d-camera}
\end{figure}

Our main goal in this paper is to reconstruct the camera positions and orientations but as the cameras are measuring  relative angles 
the best that we can hope for is a solution up to a global similarity transformation, which is often denoted metric or Euclidean reconstruction in the literature. Scale  requires extra knowledge of the world, such as common sizes of seen objects, etc. and absolute orientation 
needs much more.

\subsection{2D}

Unknowns: Each  calibrated camera, $i$, has an orientation
 parameter  and 
$2$ location parameters.
This can be encoded by a $2 \times 2$ rotation matrix $R_i$  and a $2$-dimensional
position vector $T_i$.

Each pixel, $p_{ij}$, where camera $i$ sees camera $j$ gives the equation:
\begin{equation}
p_{ij}=\frac{R_i^1  (T_j-T_i)}{R_i^2  (T_j-T_i)}
\label{eq-r-2d}
\end{equation}
where $R^k$ is the $k^{th}$ row of $R$.
$$
p_{ij}{R_i^2  (T_j-T_i)}={R_i^1  (T_j-T_i)}
$$
which can be expressed in angular terms as:
$$
\sigma_{ij}=\arctan{p_{ij}}
$$

There is an  ambiguity
in both \eqref{eq-r-2d} and the above definition,
as $R(\theta) = -R(\theta+\pi)$ thus seeing things only in front of the camera must be also be enforced/checked.

A 2D similarity transformation has 4 parameters and the following table gives the equation/parameter counts for a small number of cameras.

\begin{table}[h!]
\centering
\begin{tabularx}{\columnwidth}{|X|X|X|X|} 
 \hline\
 cameras&parameters up to a global similarity&minimal \#  of camera sightings required & possible \# of sightings /pixels\\
 \hline
 n&3n-4&3n-4&n(n-1)\\
 \hline
 2 & 2 &2&2  \\ 
3 & 5 & 5 &6\\ 
4& 8& 8&12\\
5& 11&11&20 \\
6&14&14&30\\
 \hline
\end{tabularx}
\vspace{.1cm}
\caption{Minimal number of sightings required for full reconstruction up to a global similarity transformation in 2D.}
\label{table:2}
\end{table}

\subsection{3D}

In 3D each $R$ has 3 degrees of freedom and each $T$ has 3 degrees of freedom.
Each pixel gives 2 equations:

$$
p_x=\frac{R_i^1  (T_j-T_i)}{R_i^3  (T_j-T_i)}
~~~~~~p_y=\frac{R_i^2  (T_j-T_i)}{R_i^3  (T_j-T_i)}
$$
or
$$
p_x {R_i^3  (T_j-T_i)}={R_i^1  (T_j-T_i)}
$$
$$
p_y {R_i^3  (T_j-T_i)}={R_i^2  (T_j-T_i)}
$$
A 3D similarity transformation has 7 parameters.

\begin{table}[h!]
\centering
\begin{tabularx}{\columnwidth}{|X|X|X|X|} 
 \hline\
 cameras&parameters up to a global similarity&minimal \#  of camera sightings required & possible \# of sightings\\
  \hline
 n&6n-7&3n-3&n(n-1)\\
 \hline
 2 & 5 & &2 \\ 
3 & 11 &6 &6 \\ 
4& 17& 9&12\\
5& 23&12&20 \\
6&29&15&30\\
 \hline
\end{tabularx}
\vspace{1mm}
\caption{Lower bound on the number of sightings  for a full reconstruction up to a global 3D similarity transformation}
\label{table:1}
\end{table}

\section{Recovering the orientations of the cameras}

\begin{figure}[h]
\begin{center}\begin{tikzpicture}
\tikzset{vertex/.style = {shape=circle,draw,minimum size=1.5em}}
\tikzset{edge/.style = {->,> = latex'}}
\node[vertex] (1) at  (0,1) {$i$};
\node[vertex] (2) at  (2,1) {$j$};
\draw[edge] (1) to node[auto,swap] {(a)}  (2);
\draw[edge] (2) to (1);
\end{tikzpicture}  
\hspace{3mm}
\begin{tikzpicture}
\tikzset{vertex/.style = {shape=circle,draw,minimum size=1.5em}}
\tikzset{edge/.style = {->,> = latex'}}
\node[vertex] (1) at  (2,2.5) {$i$};
\node[vertex] (2) at  (1,1) {$j$};
\node[vertex] (3) at  (3,1) {$k$};
\draw[edge] (1) to (2);
\draw[edge] (2) to (1);
\draw[edge] (1) to (3);
\draw[edge] (3) to (1);
\draw[edge] (3) to (2);
\draw[edge] (2) to node[auto,swap] {(b)} (3);
\end{tikzpicture}
\begin{tikzpicture}
\tikzset{vertex/.style = {shape=circle,draw,minimum size=1.5em}}
\tikzset{edge/.style = {->,> = latex'}}
\node[vertex] (1) at  (2,2.5) {$i$};
\node[vertex] (2) at  (1,1) {$j$};
\node[vertex] (3) at  (3,1) {$k$};
\draw[edge] (1) to (2);
\draw[edge] (2) to (1);
\draw[edge] (1) to (3);
\draw[edge] (3) to (1);
\draw[edge] (2) to node[auto,swap] {(c)} (3);
\end{tikzpicture}
\caption{\small Three typical configurations with linear solutions
associated with equation \eqref{duo}: (a) double bidirectional,   (b) a triple bidirectional loop and (c) the minimal 3 camera setup}
\label{graphs11}
\end{center}
\end{figure}
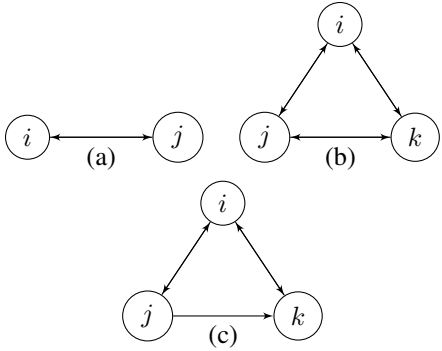

\noindent For the 2D case we  use a complex encoding of the camera parameters.
Let there be  2D cameras at location $z_j$ 
($z \in \mathbb{C}$, complex)
and
oriented at angles $\phi_j$ in the plane.
Denote  the normalized vector between two  points
$$\displaystyle u_{jk} =\frac{z_k - z_j}{|z_k - z_j|}\in \mathbb{S}^1$$
and the orientation of the $j$-th camera  by $e^{i\phi_j}$. When the $j$-th camera ``sees'' the $k$-th one we have
\begin{equation}
 e^{i\phi_j}  u_{jk} = v_{jk}
 \label{cam}
\end{equation}
where $v_{ij}\in \mathbb{S}^1$ 
is a unit vector in the direction of the pixel $p_{ij}$ in the $i$-th camera's coordinate system, see Figure \ref{2d-camera}.
With this normalization, we have a relation on the unit circle which  reduces to  equality of the arguments, namely
\begin{equation}\label{arg}
  \phi_i + \alpha_{ij} = \sigma_{ij}
\end{equation}
where  $\alpha_{ij} = \arg{u_{ij}}$ and $\sigma_{ij} = \arg{v_{ij}}$.
Since obviously $u_{ji} = -u_{ij}$, i.e., $\alpha_{ji} = \alpha_{ij}+\pi$, one may derive from \eqref{arg} in the case of mutual viewing
of $i$ and $j$, $i \leftrightarrow j$ illustrated in Figure~\ref{graphs11}:a, 
the following relation:
  \begin{equation}\label{duo}
 \phi_i -\phi_j = \sigma_{ji}- \sigma_{ij} + \pi
\end{equation}
where the $\sigma_{ij}$'s are known.
This linear system, Equation~\ref{duo}, in the $\phi_i$'s resolves the orientation problem for a connected component of  bidirectional links (mutual views). Note that  angles are equivalent mod $2\pi$.
This follows as there are at least $m-1$ bidirectional links for the $m$ cameras and there are $m-1$ unknown $\phi_i$'s, as one of them is free due to the global similarity ambiguity.

From Equation \eqref{cam} we  can get the relative difference vectors, $\alpha_{ij}$.
In this way 2 or 3 mutually seeing cameras can be completely solved (up to scale). 

In case of Figure~\ref{graphs11}:b, a triangle,
the angles at vertices are straightforward to compute, for example, the angle at  $i$ is computed from the two cameras it sees, $p_{ij}$ and $p_{ik}$
\begin{equation}\label{trio}
\measuredangle(jik) =  \sigma_{ij}-\sigma_{ik}.
\end{equation}
The 3 angles fix a triangle up to similarity and using Equation \eqref{cam}  the orientations can be computed.
Moreover, one of the links can be uni-directional, Figure~\ref{graphs11}:c,
as 2 angles determine the third, this is the minimum number of edges or pixels seen, Table ~\ref{table:2}, for a full 3 camera reconstruction.

\section{The 3D Setting}

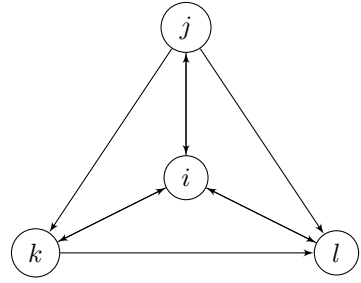
\begin{figure}
 \begin{center}
\begin{tikzpicture}
\tikzset{vertex/.style = {shape=circle,draw,minimum size=1.5em}}
\tikzset{edge/.style = {->,> = latex'}}
\node[vertex] (0) at  (4,1) {$i$};
\node[vertex] (1) at  (4,3) {$j$};
\node[vertex] (2) at  (2,0) {$k$};
\node[vertex] (3) at  (6,0) {$l$};
\draw[edge] (1) to (0);
\draw[edge] (2) to (0);
\draw[edge] (3) to (0);
\draw[edge] (0) to (1);
\draw[edge] (0) to (2);
\draw[edge] (0) to (3);
\draw[edge] (1) to (2);
\draw[edge] (1) to (3);
\draw[edge] (2) to (3);
\end{tikzpicture}
\caption{the tetrahedral configuration in 3D }
\label{graph3d1}
\end{center}
\end{figure}

In case of Figure~\ref{graphs11}:b, a triangle, the 3D case is the same as the 2D case. 
Given a tetrahedron,  Figure~\ref{graph3d1}, with bidirectional links
each of the 4 (although 3 is sufficient) triangular faces can be computed and together they form the
vertices of the tetrahedron. Each camera sees the other 3 so that the orientations can be computed using methods of,
\cite{Faugeras:1986:RRL:9356.9358,Arun:1987:LFT:28809.28821}, either using quaternions or rotation matrices and svd.

We present a more explicit solution  in the 3D case as it is more complicated: on the one hand, the camera rotations $R_i$ are  representatives of the group ${\rm SO}(3)$, which is not commutative and depends on three (not just one) parameters, while  the mutual orientations ${\bf u}_{ij}$ inhabit the unit sphere $\mathbb{S}^2$ rather than the unit circle, so we need to dispose of the convenience of a complex representation. Nevertheless, one may use a similar idea to isolate them from the general equation and obtain a relation analogous to \eqref{duo}. To begin with, just like in the 2D case, we make use of normalized relative position vectors
$$\displaystyle {\bf u}_{jk} =\frac{{\bf T}_k - {\bf T}_j}{||{\bf T}_k - {\bf T}_j||}\in \mathbb{S}^2$$
where  ${\bf T}_i$'s is the location  of camera $i$.
The image coordinates ${\bf v}_{ij}$ are also unit vectors. This yields an equation on $\mathbb{S}^2$ similar to \eqref{cam} in the form
\begin{equation}\label{cam3}
  R_i {\bf u}_{ij} = {\bf v}_{ij}.
\end{equation}
Note that the obvious relation ${\bf u}_{ij}=-{\bf u}_{ij}$ in the case of a bidirectional link yields in analogy with formula \eqref{duo}.

\begin{equation}\label{3d}
 R_j R_i^{-1}:\quad  {\bf v}_{ij}\; \rightarrow \;-{\bf v}_{ji}
\end{equation}

Due to the increased complexity in the three-dimensional case one cannot provide a straightforward analogy to formula \eqref{trio}, however, triangulation (up to global similarity) is still possible on a certain  graphs. 
Let $\Gamma_0$ denote the complete  bidirectional graph (with at least three vertices) containing a reference camera with given position $T_0$ and orientation $R_0$.
To obtain the camera orientations explicitly we  use a convenient parametrization of $\mathrm{SO}(3)$ due to Rodrigues (see \cite{biv} for details), which uses projective quaternion vector-like coordinates instead of matrices, i.e.,
\begin{equation}\nonumber
  {\bf c} = \frac{\langle q \rangle_2}{\langle q \rangle_0}\in\mathrm{SO}(3),\qquad q\in \mathbb{H}^\times
\end{equation}
where $\langle \cdot \rangle_k$ denotes grade projection (in this case bivector and scalar component) and $\mathbb{H}^\times$ the group of invertible (i.e., nonzero) quaternions. Quaternion multiplication then yields the group composition law in the simple form
\begin{equation} \label{comp}
\langle\, {\bf c}_2, {\bf c}_1 \rangle = \frac{{\bf c}_2+{\bf
c}_1+{\bf c}_2\times{\bf c}_1}{1-{\bf c}_2 \cdot{\bf c}_1}
\end{equation}
with trivial and inverse elements given respectively as
\begin{equation}\nonumber
\langle  {\,\bf c},\, 0 \,\rangle = \langle\, 0, \,{\bf c}\,\rangle ={\bf c},\qquad \qquad \langle \, {\bf c},\, -{\bf c} \,\rangle = 0
\end{equation}

The link to the usual matrix representation is provided by the Cayley transform
\begin{equation}\label{cay}
R({\bf c})=\frac{\mathcal{I}+{\bf c}^\times}{\mathcal{I}-{\bf c}^\times}=
\frac{(1-{\bf c}^2)\,\mathcal{I}+2\,{\bf c}{\bf c}^t+2\,{\bf c}^\times}{1+{\bf c}^2}
\end{equation}
where $\mathcal{I}$ stands for the identity transformation and $\_^\times$ denotes the Hodge dual used in the cross product, i.e., ${\bf c}^\times {\bf a} = {\bf c}\times {\bf a}$. Next, based on a remark by Pi\~{n}a \cite{pin2}  if a rotation $R({\bf c})$ sends ${\bf u}$ to ${\bf v}\in\mathbb{S}^2$ its axis belongs to the bisector plane spanned by ${\bf u}+{\bf v}$ and ${\bf u}\times {\bf v}$ and its Rodrigues' vectorial parameter may  be expressed as\footnote{with the exception of the case ${\bf u}=-{\bf v}$, in which ${\bf c}$ is infinite in magnitude and oriented arbitrarily in the plane ${\bf u}^\perp$.}
\begin{equation}\label{vp}
  {\bf c} = \frac{ {\bf u}\times {\bf v} + \lambda({\bf u} + {\bf v})}{1 + {\bf u}\cdot {\bf v}}
\end{equation}
where $\lambda\in \mathbb{RP}^1$, it can be  $\infty$, is an undetermined parameter. Note that each bidirectional link in the graph gives an equation in the form \eqref{3d} and yields one such parameter.
It is possible in principle to follow the analogy with the 2D case constructing a system of such links (e.g. a tetrahedron). Before doing that, however, due to the increased complexity (the corresponding system would be non-linear) we prefer to  simplify our reference camera  in advance. For instance, if $R_i$ is known, then from \eqref{cam3} one has
$${\bf u}_{ij} = - {\bf u}_{ji} =R_i^{-1}{\bf v}_{ij}= {\bf v}_{ij}'$$
hence, from formula \eqref{vp} we have
\begin{equation}\label{cj}
  {\bf c}_j = \frac{{\bf v}_{ji}\times{\bf v}_{ij}' + \lambda_j({\bf v}_{ij}'-{\bf v}_{ji})}{1-{\bf v}_{ji}\cdot{\bf v}_{ij}'}\cdot
\end{equation}
Next, we assume to have a triangular graph with only bidirectional links (Figure \ref{graphs11}c) with the reference camera at one of the vertices. This immediately yields an additional relation \eqref{3d} for the link $j\leftrightarrow k$ where $R_i$ and $R_j$ may be described by means of their Rodrigues' parameters ${\bf c}_{i,j}$ using formula \eqref{cj}. This extra link, on the other hand, yields a relation of the type \eqref{vp} for the composition $ \langle {\bf c}_k, -{\bf c}_j \rangle$, namely the 3 component vector equation:
\begin{equation}\label{expl}
   \frac{{\bf c}_k  -{\bf c}_j+ {\bf c}_j\times {\bf c}_k}{1+{\bf c}_j\cdot{\bf c}_k} = \frac{{\bf v}_{kj}\times{\bf v}_{jk} + \lambda_{jk}({\bf v}_{jk}-{\bf v}_{kj})}{1-{\bf v}_{jk}\cdot{\bf v}_{kj}}\cdot
\end{equation}

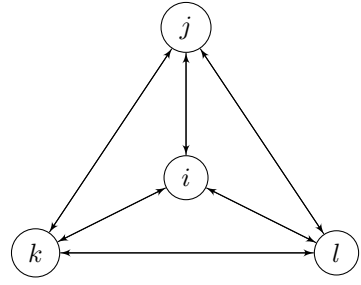
\begin{figure}
 \begin{center}
\begin{tikzpicture}
\tikzset{vertex/.style = {shape=circle,draw,minimum size=1.5em}}
\tikzset{edge/.style = {->,> = latex'}}
\node[vertex] (0) at  (4,1) {$i$};
\node[vertex] (1) at  (4,3) {$j$};
\node[vertex] (2) at  (2,0) {$k$};
\node[vertex] (3) at  (6,0) {$l$};
\draw[edge] (1) to (0);
\draw[edge] (2) to (0);
\draw[edge] (3) to (0);
\draw[edge] (0) to (1);
\draw[edge] (0) to (2);
\draw[edge] (0) to (3);
\draw[edge] (1) to (2);
\draw[edge] (2) to (1);
\draw[edge] (1) to (3);
\draw[edge] (3) to (1);
\draw[edge] (3) to (2);
\draw[edge] (2) to (3);
\end{tikzpicture}\caption{\small the bidirectional tetrahedral configuration in 3D which yields an over-determined system of quadratic equations \eqref{hor} assuming the reference camera is placed at one of the vertices.}
\label{graph3d}
\end{center}
\end{figure}

\noindent To avoid dealing with the additional unknown parameter $\lambda_{jk}$, we project \eqref{expl} to a subspace orthogonal to the vector $ {\bf v}_{[j,k]} = {\bf v}_{jk} - {\bf v}_{kj}$ as long as the latter is nonzero. More precisely, in the regular setting ${\bf v}_{kj}\times {\bf v}_{jk}\neq 0$, by dot-multiplying the above vector relation with two conveniently chosen orthogonal vectors in  $ {\bf v}_{[j,k]}^\perp$  we obtain a pair of quadratic equations for $\lambda_j$ and $\lambda_k$ in the form

\begin{eqnarray}\nonumber
 \langle {\bf c}_k, -{\bf c}_j \rangle\cdot\left({\bf v}_{kj}\times {\bf v}_{jk}\right) & = & 1 + {\bf v}_{jk}\cdot {\bf v}_{kj}  \\[-1.5mm]
 & \label{hor}\\[-1.5mm]
 \langle {\bf c}_k, -{\bf c}_j \rangle \cdot\left({\bf v}_{kj} + {\bf v}_{jk}\right) & = & 0.\nonumber
\end{eqnarray}
On the other hand, since the $ {\bf v}_{jk}$'s are normalized, ${\bf v}_{kj}\times {\bf v}_{jk}=0$ yields ${\bf v}_{jk} = \pm {\bf v}_{kj}$. In the former case (positive sign) $\langle {\bf c}_k, -{\bf c}_j \rangle$ is a half-turn whose axis is oriented arbitrarily in the plane orthogonal to ${\bf v}_{jk}$, as we already pointed out. Thus, the first equation in \eqref{hor} is replaced with the condition that the denominator vanishes (Rodrigues' parametrization associates half-turns with the plane at infinity), so one still can resolve
\begin{eqnarray}\nonumber
\left({\bf c}_k  -{\bf c}_j+ {\bf c}_j\!\times {\bf c}_k\right) \cdot{\bf v}_{jk} = 1+{\bf c}_j\cdot{\bf c}_k  = 0
\end{eqnarray}
while ${\bf v}_{jk} = - {\bf v}_{kj}$ means that $R_k R_j^{-1} = \mathcal{I}$, hence $\langle {\bf c}_k, -{\bf c}_j \rangle = 0$. The latter yields ${\bf c}_k = {\bf c}_j =0 $ only in the case ${\bf c}_k\times {\bf c}_j \neq 0$, otherwise we end up with a one-parameter set of solutions satisfying ${\bf c}_j= {\bf c}_k $. In order to eliminate this indeterminacy, one needs additional links, e.g. a tetrahedron as shown in Figure \ref{graph3d}.

\section{Adding a single  camera}
One goal is to compute the locations and orientations of all the cameras.
The simple strategy of solving it all at one time using a nonlinear optimization procedure can  get stuck in a non-optimal solution. A feasible strategy is  to serially calculate  the structure, where each new camera $p$ is connected with at least two vertices in $K$, the already solved subset of cameras. 

If we can continue this process this gives a sequential algorithm. $K$ will always, partially arbitrarily due to the similarity freedom, assumed to be completely fixed with no free degrees of freedom.

\subsection{2D}

Let $|K|\ge 2$ be a set of cameras which are completely determined in the plane. To add one camera, $p$, requires at least
three sightings between $K$ and $p$, as $p$ has 3 unknown parameters.

Any three sightings 
between $K$ and $p$ with at least one from $p$ to $K$, otherwise $p$'s orientation cannot be computed, works. The cameras in $K$ are considered as landmarks. 
\begin{figure}[h]
   \centering
\includegraphics[width=\columnwidth]{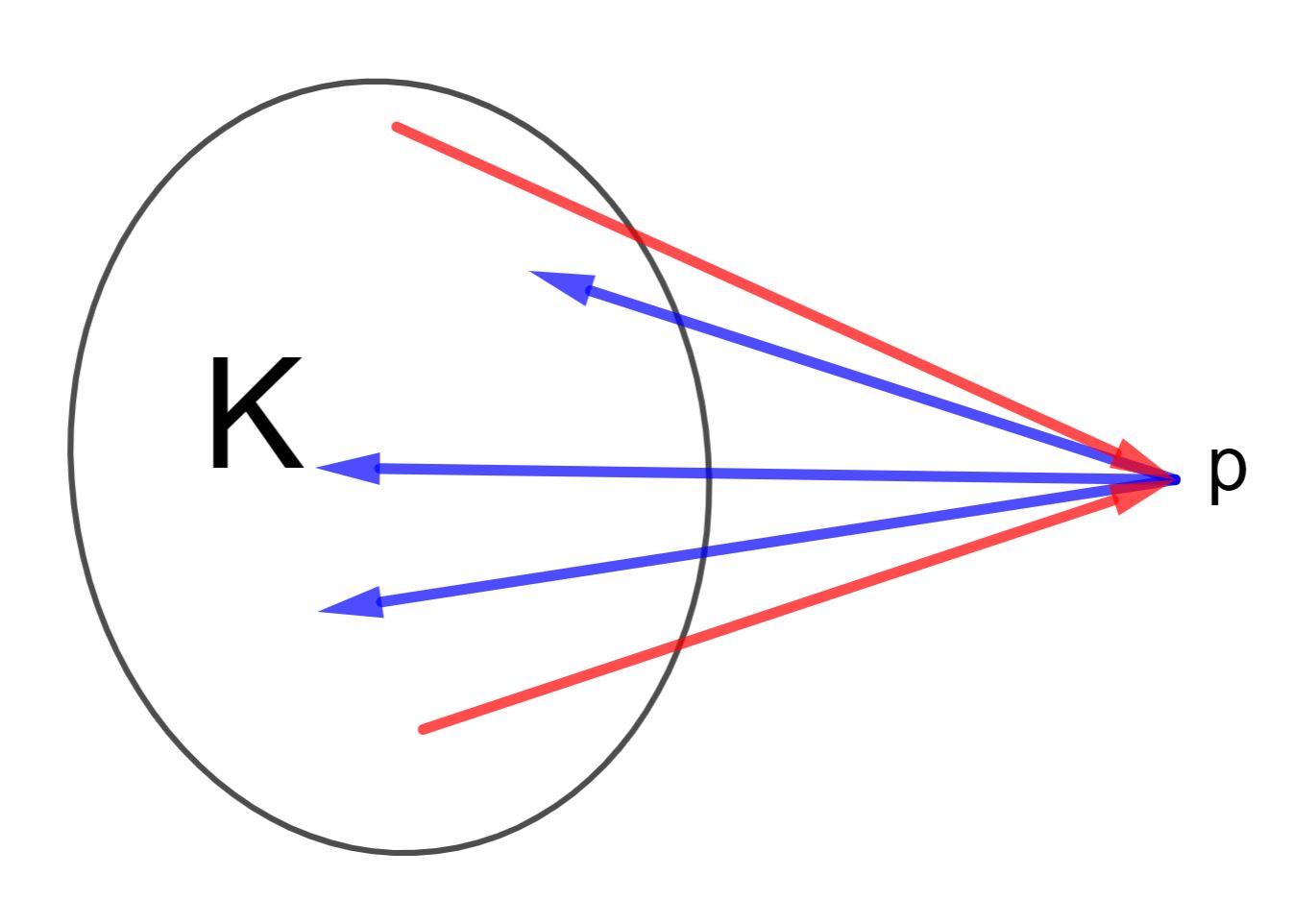}
\caption{When the cameras in $K$ are known,  computing $p$ requires the pixels from any three  arrows including at least 1 blue one for the 2D case and 2 blue ones for the 3D case.}
\end{figure}

\begin{figure}[h]
    \centering
    \includegraphics[width=0.4\columnwidth]{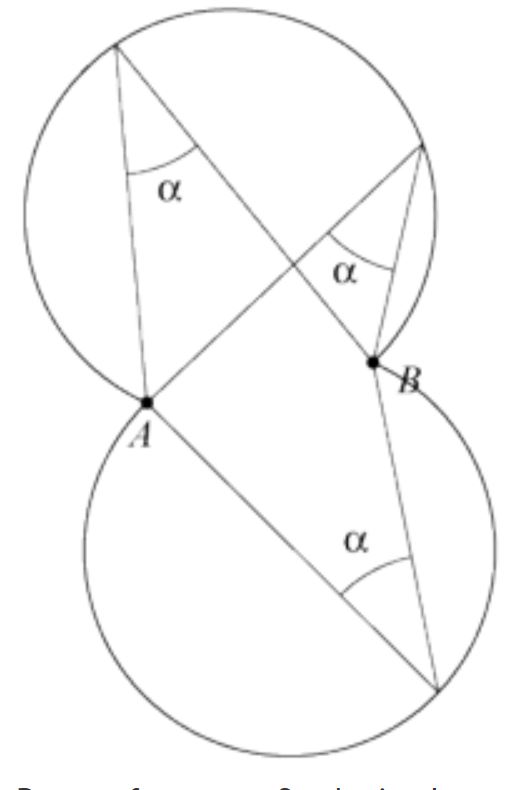}
     \includegraphics[width=0.5\columnwidth]{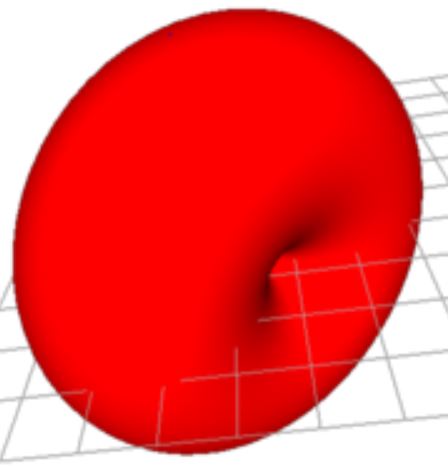}
    \caption{Locus of points viewing a segment $\overline{AB}$
    with a constant angle in 2 and 3 dimensions}
    \label{locus-2}
\end{figure}

When there are 2 sightings from $p$ to $K$, $p$ is on the locus of points seeing a segment under a constant angle, two circular arcs, see Figure \ref{locus-2}. When there are 2 sightings from $K$ to $p$ finding the location of $p$ is known as triangulation. The third sighting 
determines $p$, at least up to a finite ambiguity.

Of course, at the end of the process and even during it a global bundle adjustment should be carried out.

\subsection{3D}

Let $|K|\ge 2$ be a set of cameras which are completely determined in the plane. To add one camera, $p$, requires at least
three sightings between $K$ and $p$, as $p$ has 6 unknown parameters and each pixel gives 2 equations.

Any three sightings between $K$ and $p$ with at least 2 from $p$ to $K$, otherwise the orientation (3dof) of $p$ is under determined, works. The cameras in $K$ are considered as landmarks. 

When there are 2 sightings from $p$ to $K$, $p$ is on the locus of points seeing a segment under a constant angle, a bialli, see Figure \ref{locus-2}. When there are at least 3 sightings from $p$ to $K$ this is known as exterior parameter calibration or pose estimation.
The third sighting 
determines $p$,  up to a finite (at most 4 solutions) ambiguity \cite{MVG,en2018rpnet,Cao2018}.
When there are 2 sightings from $K$ to $p$ finding the location of $p$ is known as triangulation. Two more sightings from $p$ to $K$  determine $p$'s orientation.

\section{When can all the cameras see each other?}

\begin{figure}[h]
\centering
\begin{tikzpicture}
\pie[ text = inside, pos ={0,0} , sum =360 , after number =, radius=1,rotate =-2,/tikz/nodes={text opacity=0,overlay}]{95/1 }
\pie[ text = inside,pos ={4 ,0} , sum =360 , after number =, radius=1,rotate =88]{95/2 }
\pie[ text = inside,pos ={4 ,4} , sum =360 , after number =, radius=1,rotate =178]{95/ 3}
\pie[ text = inside,pos ={0 ,4} , sum =360 , after number =, radius=1,rotate =268]{95/4 }
\end{tikzpicture}
\caption{Each camera can see all the others}
\label{convex}
\end{figure}
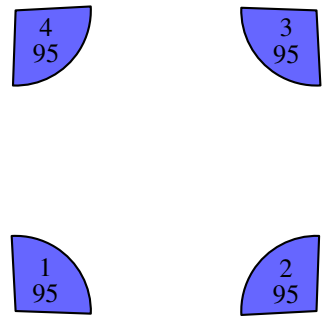

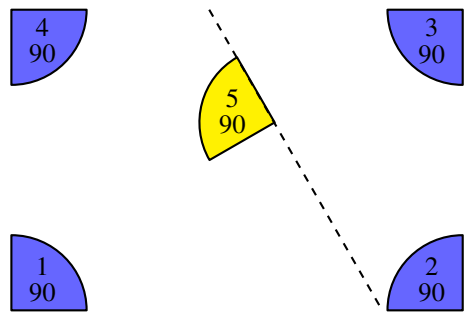
\begin{figure}
\centering
\begin{tikzpicture}
\pie[ text = inside, pos ={0,0} , sum =360 , after number =, radius=1,rotate =0,/tikz/nodes={text opacity=0,overlay}]{90/1 }
\pie[ text = inside,pos ={6 ,0} , sum =360 , after number =, radius=1,rotate =90]{90/2 }
\pie[ text = inside,pos ={6 ,4} , sum =360 , after number =, radius=1,rotate =180]{90/ 3}
\pie[ text = inside,pos ={0 ,4} , sum =360 , after number =, radius=1,rotate =270]{90/4 }
\pie[ text = inside,pos ={3.5 ,2.5} , sum =360 , after number =, radius=1,rotate =120,color=yellow]{90/5 }
\draw[red,thick,dashed] (2.63,4) --  (4.93,0);
\end{tikzpicture}
\caption{The yellow camera can not see all the other ones, specifically those to the right of the  dashed line.}
\label{inner}
\end{figure}

If the cameras are not in a convex position the  one inside the convex hull cannot see everyone else, see Figure~\ref{inner}.

The number of cameras that can see each other is a monotone function of the field of view of the cameras, FOV.

\subsection{2D}
An n-gon's sum of angles   is $(n-2)\pi$
so one angle is at least  $\frac{(n-2)\pi}{n}$  giving that 
 $$ FOV \ge \frac{(n-2)\pi}{n}$$

\newdimen\R
\R=0.3cm

The regular polygons give extremal examples:
\begin{table}[h]
\centering
 {\tabulinesep=1.4mm}
\begin{tabu}{ |l|c|l|c| } 
 \hline
 vertices&FOV & FOV &\\
 \hline
 \hline
 3 & $\frac{\pi}{3}$&1.04&\begin{tikzpicture}
\draw (0:\R) \foreach \x in {120,240} {
            -- (\x:\R)
        } -- cycle (90:\R);
\end{tikzpicture} \\ 
 \hline
 4&$\frac{\pi}{2}$& 1.57&
 \begin{tikzpicture}
 \draw[xshift=2.5\R] (0:\R) \foreach \x in {90,180,...,359} {
            -- (\x:\R)
        } -- cycle (90:\R);
 \end{tikzpicture}\\
 \hline
 5&$\frac {3\pi }{5}$&1.88&
 \begin{tikzpicture}
 \draw[xshift=5.0\R] (0:\R) \foreach \x in {72,144,...,359} {
            -- (\x:\R)
        } -- cycle (90:\R);
 \end{tikzpicture}\\
 \hline
 6&$\frac{2\pi}{3}$& 2.09&
 \begin{tikzpicture}
 \draw (0:\R) \foreach \x in {60,120,...,359} {
                -- (\x:\R)
            }-- cycle (90:\R);
 \end{tikzpicture}\\
 \hline
 7&$\frac{5\pi}{7}$&2.24&
 \begin{tikzpicture}
 \draw[xshift=2.5\R] (0:\R) \foreach \x in {51.4286,102.8571,...,359} {
                -- (\x:\R)
            }-- cycle (90:\R);
 \end{tikzpicture}
 \\
 \hline
 $\infty$& $\pi$&3.14&
 \begin{tikzpicture}
 \draw[xshift=2.5\R] (0:\R)  circle (\R);
 \end{tikzpicture}\\
 \hline
\end{tabu}
\vspace{1mm}
\caption{The regular polygons give extremal examples of mutually viewing cameras given their FOV (radians).}
\label{table:2d-reg}
\end{table}

\subsection{3D}
In order to dismiss degenerate solutions, the FOV in 3D will be taken as a regular pyramid. The degenerate solutions can be for example a FOV that is planar which will always be  $0$ sterdians.

 In general, finding the best distribution of $n$ cameras is a hard problem.
 Other than the platonic solids see Table \ref{table:3}
very few exact solutions are known and most solutions are computed using a numerical optimization.
A variation of this problem
"Distribution of points on the 2-sphere" is one of a list of eighteen unsolved problems in mathematics proposed by Steve Smale in 1998 \cite{smale1998mathematical}.

\begin{table}[h!]
\centering
 {\tabulinesep=1.2mm}
\begin{tabu}{ |l|c|c|c| } 
 \hline
 vertices&FOV & FOV& \\
 \hline
 \hline
 4 & $\cos ^{-1}\left({\frac {23}{27}}\right)$&0.55&
 \includegraphics[width=1.12cm]{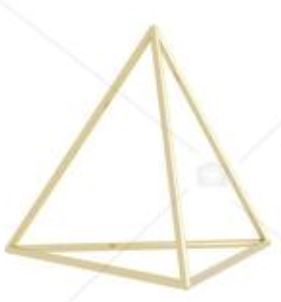}\\ 
 \hline
 6&$4\sin ^{-1}\left(\frac{1}{3}\right)$&1.35&
 \includegraphics[width=1.12cm]{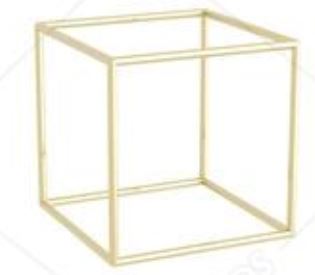}\\
 \hline
 8&$\frac {\pi }{2}$&1.57&
 \includegraphics[width=1.12cm]{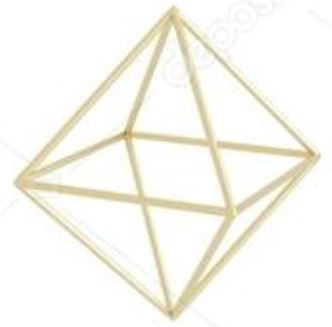}\\
 \hline
 12&$2\pi -5\sin ^{-1}\left(\frac{2}{3}\right)$& 2.63&
 \includegraphics[width=1.12cm]{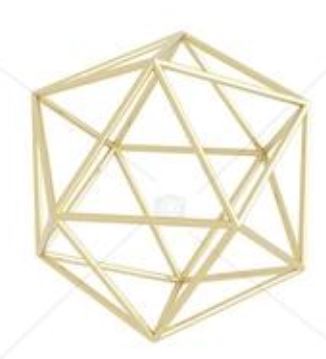}\\
 \hline
 20&$\pi- \tan ^{-1}\left(\frac{2}{11}\right)$&2.96&
 \includegraphics[width=1.12cm]{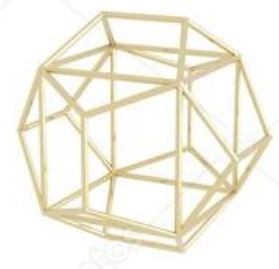}\\
 \hline
 $\infty$&$2\pi$&6.28& \includegraphics[width=1.12cm]{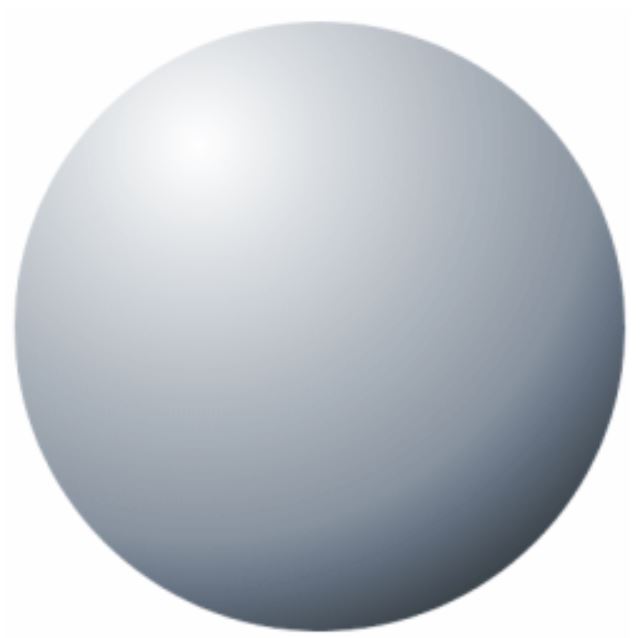}\\
 \hline
\end{tabu}
\vspace{1mm}
\caption{The platonic solids give extremal examples of number of mutually viewing cameras given their FOV (steradian).}
\label{table:3}
\end{table}

\section{Random configurations}
\subsection{Probability that a camera sees another}

\begin{figure}[h]
    \centering
    \includegraphics{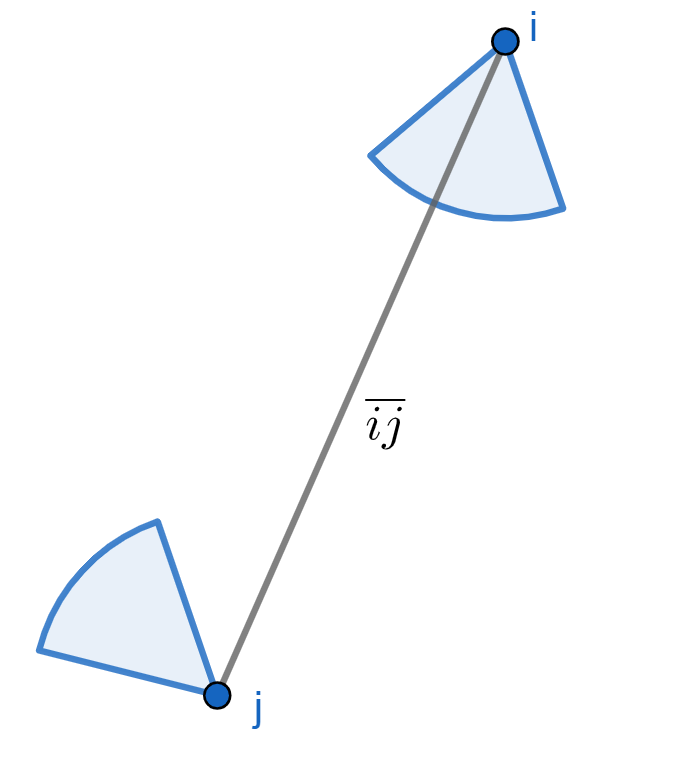}
    \caption{$i$ sees $j$,  $j$ does not see $i$}
\label{sees}
\end{figure}

The probability that one camera in a random orientation sees another, disregarding occlusions, is a monotone function of the FOV. 

When the segment between cameras $i$ and $j$,
 $\overline{ij}$,   is in the $FOV$ of $i$, $i$ sees $j$, see Figure~\ref{sees}.
 
\subsection{2D}
Assume  that a camera's orientation is uniform in $[0..2\pi]$.
\begin{equation}
   P(i ~sees~ j)=\frac{FOV}{2\pi}
 \label{fov-2pi}
\end{equation}
the expected number of the other $n-1$  cameras $i$ sees is
$$
E(\# ~of ~cameras~ i~ sees)=\frac{(n-1)FOV}{2\pi}
$$
 and 
 the expected number of sightings in the system is
$$
E(\# ~of ~cameras~ seen)=\frac{n(n-1)FOV}{2\pi}
$$

The FOV needed so that the expected number of sightings is enough for a full reconstruction:
$$
3n-4 \le \frac{n(n-1)FOV}{2\pi}
$$
or
$$
FOV \ge \frac{(3n-4){2\pi}}{n(n-1)}
$$

\begin{table}[h]
    \centering
\begin{tabu}{|l||c|c|c|c|c|c|}
\hline
 cameras&6&7&8&9&10&11\\
 \hline
 FOV&2.93&2.54&2.24&2.00&1.81&1.65\\ 
 \hline
\end{tabu}
    \caption{Minimal FOV (radians) in 2D so that in expectation there  are a sufficient number of sightings}
    \label{tab:my_label}
\end{table}

The reverse Markov inequality can be used to bound the probability, $\forall a \le n(n-1)$

$$
{\mathrm{Pr}}[ Y \leq a ] ~\le~ \frac{n(n-1)-{\mathrm E}[  Y ]}{ n(n-1)-a}. 
$$
where $Y$ is the number of sightings.

Assuming independence of the orientations of $i$ and $j$ the probability that they see each other is:
$$
\frac{FOV^2}{4\pi^2}
$$

\subsection{3D}
In 3D the same holds except the sphere's angle  is $4\pi$ steradians.

$$
P(i ~sees~ j)=\frac{FOV}{4\pi}
$$
and the expected number of the other $n-1$  cameras $i$ sees is
$$
E(\# ~of ~cameras~ i~ sees)=\frac{(n-1)FOV}{4\pi}
$$
 and 
 the expected number of sightings in the system is
$$
E(\# ~of ~cameras~ seen)=\frac{n(n-1)FOV}{4\pi}
$$

The FOV needed so that the expected number of sightings is enough for a full reconstruction:
$$
3n-3 \le \frac{n(n-1)FOV}{4\pi}
$$
or
$$
FOV \ge \frac{(3n-3){4\pi}}{n(n-1)}
$$

\begin{table}[h]
    \centering
\begin{tabu}{|l||c|c|c|c|c|c|}
\hline
 cameras&9&10&11&12&13&14\\
 \hline
 FOV&4.18&3.76&3.42&3.14&2.89&2.69\\ 
 \hline
\end{tabu}
    \caption{Minimal FOV (steradians) 3D }
    \label{tab:my_label3d}
\end{table}

Assuming independence of the orientations of $i$ and $j$ the probability that they see each other is:
$$
\frac{FOV^2}{16\pi^2}
$$

\section{Conclusion}
This paper initiated the study of 
the intra multi-view geometry of calibrated cameras  when all that they can reliably recognize is each other.   
This was carried out for both 2D and 3D cameras.

\section{Appendix: How many, randomly distributed in the sphere, cameras does each camera see?}
Here we treat the number of sighting of other cameras in a sphere as a function of distance to center, orientation and FOV.
\subsection{2D}

 In the two-dimensional case we consider a point $z_0$ in a circle of radius $r$ at distance from its center $d\in [0,r]$. It is convenient to introduce polar coordinates $\rho\in [0,r]$, $\theta\in [0,2\pi)$ choosing $z_0$ as the origin. The circle's boundary is viewed from the perspective of $z_0$ as a curve with polar equation (see \cite{li1985average} for a derivation)
\begin{equation}\label{polar}
  \rho(\theta) = d\cos{\theta}+ \sqrt{r^2-d^2\!\sin^2\!{\theta}}
\end{equation}
and it is straightforward to obtain the area of the disc segment sliced by the viewing angle $\theta\in[a,b]$ at $z_0$ as
\begin{equation}\nonumber
A = \int\limits_{a}^{b}\int\limits_0^{\rho(\theta)}{\rho\,{\rm d}\rho\,{\rm d}\theta}.
\end{equation}
Choosing a polar orientation $\phi\in [0,2\pi)$ for the camera at $z_0$ and denoting the width of the field of view $FOV = 2\delta$, one obtains the above integral as a a function of the parameters $\phi$ and $d$, or more conveniently $\epsilon = d/r$ (keeping $r$ and $\delta$ fixed), namely
\begin{equation}\label{slice}
A(\phi, \epsilon) = \frac{r^2}{2}\!\int\limits_
{\phi\!-\!\delta}^{\phi\!+\!\delta}
{\left(1\!+\epsilon^2\cos{2\theta}\!+2\epsilon\cos{\theta}\sqrt{1\!-\!\epsilon^2\sin^2\!{\theta}}\right)\!{\rm d}\theta}.
\end{equation}
The first two terms are trivial, while for the third one we use integration by parts, finally arriving at
$$\displaystyle
  A(\phi, \epsilon) = \delta r^2 + \left(\frac{\epsilon r}{2}\right)^2\sin(2\theta)
  \Big|_{\phi\!-\!\delta}^{\phi\!+\!\delta} +\qquad\qquad\qquad $$  $$\frac{r^2}{2}\left( \epsilon \sin{\theta}\sqrt{1-\epsilon^2\!\sin^2\theta}+\arcsin{\epsilon\sin{\theta}}\right)\Big|_{\phi\!-\!\delta}^{\phi\!+\!\delta}
$$
e.g. on the boundary of the unit circle $\epsilon \!= \!r\! =\! 1$ one has
$$A = 2\delta + \cos{ 2\delta} \cos {2\phi} $$
while for an arbitrarily placed camera pointing towards the center (note that we always assume $\delta  \in [0,\pi]$)
$$A = \delta + \tilde{\delta} + \sin{\tilde{\delta}}(\cos{\delta}+\cos{\tilde{\delta}}), \qquad \tilde{\delta} = \arcsin{\epsilon\sin{\delta}}.$$
Using formula \eqref{slice} one  obtains an estimate for the geometric probability
$$P = A/ A_0,\qquad A_0=\pi r^2$$
that in the case of the uniform distribution corresponds to the relative number of agents seen by the camera at $z_0$. The rotational symmetry allows us to work with the above-chosen range for the parameters $\epsilon \in [0, 1]$, $\phi\in [0,2\pi)$ and then integrate $P(\phi,\epsilon)$ dividing by $2\pi$ in order to obtain the average probability and thus, the number of agents seen by an arbitrary camera 
\begin{equation}\label{prob}
  \langle {P} \rangle = \frac{1}{2\pi^2 r^2} \int\limits_{0}^{1}\!\int\limits_{0}^{2\pi} A(\phi, \epsilon)\, {\rm d}\phi\,{\rm d}\epsilon = \frac{\delta}{\pi}=\frac{FOV}{2\pi}
\end{equation}
where we use the periodicity of the trigonometric terms in \eqref{slice} with respect to $\phi$
and get the same result as in Equation \ref{fov-2pi}. Note that since the nonzero contribution to \eqref{prob} does not depend explicitly on $\epsilon$, the above relation holds for any point on the circle. 

\subsection{3D}
 The three-dimensional setting is more complicated as we need to intersect the viewing cone of the observer
\begin{equation}\label{cone}
  x^2 + y^2 = c^2z^2, \qquad c = \tan{\delta}
\end{equation}
with a sphere of radius $r$, which in the observer's reference frame is given by the equation
\begin{equation}\label{spher}
  (x-x_0)^2 + y^2 + (z-z_0)^2 = r^2
\end{equation}
where the solid viewing angle is expressed as $FOV = 2\pi(1-\cos{\delta})$ and we use the polar symmetry to set $y_0=0$ for simplicity. Next, introducing cylindrical coordinates and assuming that the viewing cone is forward oriented, i.e., $FOV\leq 2\pi$, we  obtain the volume of the visible domain confined within the ball 
where the radial coordinate of the intersection $\rho_{\mu}$ depends on the polar angle $\varphi$ in a rather complicated way
$$  \rho_{\mu} = \frac{c}{1+c^2}\left( z_0 + cx_0\cos{\varphi} + \right. $$
$$\left.\sqrt{r^2-x_0^2 + c^2(r^2\!-\!z_0^2)+2cx_0z_0\cos{\varphi}-c^2x_0^2\sin^2{\varphi}}\right)
$$
which is greatly simplified if the cone and the sphere share a common axis of symmetry, i.e., $x_0=0$, namely
\begin{equation}\label{max}\nonumber
  \rho_{0} = \frac{c}{1+c^2}\left( z_0 + \sqrt{r^2+ c^2(r^2\!-\!z_0^2)}\right)
\end{equation}
and the above integral has an exact solution in the form
\begin{equation}\label{max1}\nonumber
V_0 = \frac{2\pi}{3}\left(r^3 -c^{-1}\rho_0^3\right) -  \frac{2\pi}{3}\left(r^2-\rho_0^2\right)^{\frac{3}{2}}+ \pi z_0 \rho_0^2. 
\end{equation}
In particular, if $FOV = 2\pi$, one ends up with $\displaystyle \rho_{0} = \sqrt{r^2-z_0^2}$ and respectively
\begin{equation}\label{max2}\nonumber
V_0 = \frac{2\pi}{3}\left(r^3  - |z_0|^3\right)  + \pi z_0 \left(r^2 - z_0^2\right).
\end{equation}
In the generic case, however, the integral cannot be resolved in elementary functions.

\end{document}